# Forecast Analysis of the COVID-19 Incidence in Lebanon: Prediction of Future Epidemiological Trends to Plan More Effective Control Programs

Salah El Falou[1], Fouad Trad[2*]
[1]Faculty of Science, Lebanese University, Ras Maska – Al Koura, Lebanon
[2]Faculty of Engineering, Lebanese University, Ras Maska – Al Koura, Lebanon

*Abstract*

**Ever since the COVID-19 pandemic started, all the governments have been trying to limit its effects on their citizens and countries. This pandemic was harsh on different levels for almost all populations worldwide and this is what drove researchers and scientists to get involved and work on several kinds of simulations to get a better insight into this virus and be able to stop it the earliest possible.**
**In this study, we simulate the spread of COVID-19 in Lebanon using an Agent-Based Model where people are modeled as agents that have specific characteristics and behaviors determined from statistical distributions using Monte Carlo Algorithm. These agents can go into the world, interact with each other, and thus, infect each other. This is how the virus spreads. During the simulation, we can introduce different Non-Pharmaceutical Interventions - or more commonly NPIs - that aim to limit the spread of the virus (wearing a mask, closing locations, etc).**
**Our Simulator was first validated on concepts (e.g. Flattening the Curve and Second Wave scenario), and then it was applied on the case of Lebanon. We studied the effect of opening schools and universities on the pandemic situation in the country since the Lebanese Ministry of Education is planning to do so progressively, starting from 21 April 2021.**
**Based on the results we obtained, we conclude that it would be better to delay the school openings while the vaccination campaign is still slow in the country.**

## I- Introduction

Coronavirus Disease 2019 – mainly referred to as COVID19 – is a highly infectious disease caused by severe acute respiratory syndrome coronavirus 2 (SARS-CoV-2). This disease originated in Wuhan, Hubei Province, China, in December 2019 [1], then its spread escalated and caused an outbreak in almost every country on the globe. Since its beginning, governments are struggling to find different ways to reduce its impact on their countries. With the absence of vaccines at the early stages of the pandemic because of the newness of this virus back then, a lot of Non-Pharmaceutical Interventions (NPIs) [2] have been taken to limit its spread as much as possible. These NPIs range from universal mask-wearing and social distancing to global lockdowns. Their levels varied depending on the severity of the situation in different countries (defined by the daily number of infections or deaths for example).

But how can governments choose the right action at the right time? Here is where simulations can help them most.

A lot of effort has been put to simulate the spread of COVID-19 using different modeling techniques [3]. Some of those techniques rely on pure signal processing and time series analysis [4, 5]. The problem with those techniques is that they just take the curve and try to predict based on a trend how the pandemic will progress, so it only takes previous observations to predict based on them. Others rely on solving differential equations that characterize a certain model [6–8]. The most famous across those models are the SIR (Susceptible - Infected - Recovered) and the SEIR (Susceptible - Exposed - Infected - Recovered) models. The problem with this technique is that it does not take into consideration the complex human behavior that plays an important role in the spread of the virus: they consider the population to be homogeneous which is not always the case. This is where the final modeling technique comes to the rescue: It is Agent-based modeling [9–11]. In this technique, individuals are modeled as agents having specific needs and characteristics, and thus, are singular by the actions that they do. This is how we can study human behavior's effect on the pandemic situation.

In this study, we built an Agent-Based model that simulates the spread of the virus in a virtual country where agents interact with each other. We then tested this simulator on a virtual country similar to Lebanon in its aspects and saw the effect of all previous governmental interventions, and we obtained a pandemic curve very similar to the real one. After that, and based on the willingness of the Lebanese Education Ministry to open schools again progressively starting from 21 April 2021, until 17 May 2021 [12], we simulated the effect of such an action on the pandemic evolution, taking into account the delay in the vaccination process in the country.

So, our outline will be as follows: we will start by introducing COVID-19. After that, we will talk about the agent-based modeling highlighting its relevance for this problem, and then, we'll talk about the simulator that we built and validated on concepts of COVID-19 (Flattening the curve and Second Wave Scenario) and on real data (the Lebanese Case). In the end, we use the tuned model to forecast the effect of future actions (Opening Schools and Universities) in Lebanon.

---

[*] **Corresponding Author**: `fouad.trad@ul.edu.lb`, `fouadtrad98@gmail.com`

## II- COVID-19

COVID-19 is a highly transmissible disease. On March 11, 2020, the World Health Organization (WHO) declared it to be a global pandemic, and by 21 April 2021, there have been 141,754,944 confirmed cases, including 3,025,835 deaths [13]. In Lebanon, the first case was registered on February 21, 2020 [14], and by 29 March 2021, there have been 511,398 reported cases including 6,959 deaths according to WHO [15]. By 26 April 2021, 4% of the Lebanese population have taken the first dose of the vaccine, and 2.2% have been fully vaccinated [16].

COVID-19 is caused by a strain of Coronavirus, namely, Severe acute respiratory syndrome coronavirus 2 (SARS-CoV-2), and is highly infectious. While originating in Wuhan, China in December 2019, the disease has spread to most of the world's nations. Countries have taken restrictive measures (i.e., lock-down) to contain the outbreak of the virus. This disease has heavily affected the economies of most countries leading to the global coronavirus recession or the Great Shutdown [17].

SARS-CoV-2 is spread when a person who is susceptible to infection inhale droplets coming from an infected person through coughing or sneezing.

The nature of this infection has led to the emergence of various preventive practices that are considered intrusive such as: Wearing masks, washing hands, staying at home, etc.

## III- Agent-Based Modeling

Agent-Based Modeling is a go-to when we need to model complex dynamic systems [18]. This modeling technique is considered a "from the ground up" approach [19] where agents interact with each other and based on that, we get the results that we're looking for without explicitly programming them (we only need to program the agents' behavior).

This approach takes then into consideration several things that were not taken in other studies (where the population is homogeneous). Each agent has its characteristics and behaviors [20] that would affect the spread of the virus in a certain way, and that's what happens in reality: individual people contributing in a way or another to the spread of the virus due to their different behaviors, and this is what makes the simulation process more realistic.

More detailed information and simulation examples about Agent-Based Modeling can be found in [21–23].

## IV- Simulator Overview

The goal of our simulator is to forecast the evolution of COVID-19 inside of a country with the presence of different control measures that can be applied inside of it. The pandemic situation is directly dependent on the agent's behavior inside of the country, the places where he goes, and the people he interacts with. The agent's behavior on the other hand is affected by the control strategies imposed inside of the country. The good thing about the simulator is that it can give an idea about how the pandemic situation can become after imposing a virtual control measure inside of the country. This way governments can benefit from it while seeing the effect of their actions before doing them, and then adapt the best measures to limit the virus spread.

The three key components inside of our simulator are the Virus that will spread inside of a country, the Country itself, and the people (or agents) living inside of it. We will discuss each of them in the next sections.

### A. The Virus

The virus that we want to simulate in our study is the SARS-CoV-2, but other viruses can also be modeled simply by changing the parameters. Those parameters are:

***The Latent Period:*** It is the period that lies between the time when a person becomes infected (or exposed), and the time he becomes infectious (or can infect others). The Coronavirus is defined by a latent period of 3 days [24].

***The Incubation Period:*** It is the period that lies between the time when a person is infected, and the start of his symptoms to appear. The Coronavirus is defined by an incubation period of 5 days [25].

***The Infectious Period:*** The time during which a person can infect others. After the start of the symptoms, a person can infect the others for 10 days in most of the cases according to CDC [26]. So, the infectious period in the Coronavirus starts 2 days before symptoms develop (when the latent period finishes) and it ends 10 days after symptoms start.

***Symptoms:*** Each virus has specific symptoms. The symptoms of COVID-19 include coughing, fatigue, fever, etc [27]. These symptoms can be categorized between Mild/Moderate, Severe, or Critical. A person can also be asymptomatic (does not develop symptoms). Usually, children and adolescents are more likely to be asymptomatic compared to adults [28, 29]. After developing a symptom, a person can recover or die. Death and recovery rates vary depending on the severity of the symptom: Asymptomatic and Mild cases always recover, but Severe and Critical cases have a death rate of 15% and 50% respectively [30]. If the person has severe or critical symptoms, he will require hospital care after 9, and 10 days respectively of his infection [30]. Since our goal is to simulate the case of Lebanon, the percentage of people that develop specific kinds of symptoms is taken from the daily reports of the Ministry of Public Health (MOPH) in Lebanon in the COVID-19 case distribution section [31].

All of the characteristics and the numbers related to this virus are represented in *Fig. 1*.

## B. The Country

For our aim to simulate the spread of the virus in Lebanon, we build a small country that represents Lebanon in several aspects. The country is defined by several characteristics that can be summarized as follows:

*Locations:* Each country can have several locations or places that the agent can visit. The country that we built has houses, hospitals, markets, malls, nightclubs, companies, churches, mosques, universities, schools, and an airport. Each location has a specific opening and closing time on each day of the week (except houses of course).

*House Population Distribution:* This is a statistical distribution. It reflects the percentages of houses having a specific number of people living inside of them. For example, in Lebanon, 20% of the houses contain 4 people living in them, 10% contain 5, etc. The house population distribution for our country is based on the Central Admissions of Statistics for Population in Lebanon (CAS)[32] and can be found in *Table 1*.

Table 1. House Population Distribution in Lebanon

| Number of People Living in the house | Percentage |
|---|---|
| 1 | 10% |
| 2 | 18% |
| 3 | 18% |
| 4 | 20% |
| 5 | 16% |
| 6 | 10% |
| 7 | 4% |
| 8 | 4% |

*Age Distribution:* This is another statistical distribution. It shows us the percentage of people having a specific age range inside of a country. For example, in Lebanon, 16% of the population has an age lying between 10 and 19 years old.

This distribution is also based on the CAS for Population in Lebanon [32] and can be found in *Table 2*

Table 2. Age Distribution in Lebanon

| Age Group | Percentage |
|---|---|
| 0 - 4 | 8% |
| 5 - 9 | 8% |
| 10 - 19 | 16% |
| 20 - 29 | 17% |
| 30 - 39 | 13% |
| 40 - 49 | 11% |
| 50 - 59 | 11% |
| 60 - 69 | 8% |
| 70 - 79 | 5% |
| 80 - 89 | 3% |

*Religion Distribution:* This distribution reflects the percentage of people having a certain religion. For example, in Lebanon, we have 38% of the population that is Christian, 57% that are Muslim, and 5% that have other religions [33]

## C. The Agent

The agent is defined by several characteristics when created at the beginning of the simulation:

*Age:* The first thing an agent gets is his age. This is the most significant factor for an agent because it will define its other characteristics. The age of the agent is sampled from the Age distribution in the country using Monte Carlo Algorithm.

*Profession:* Each agent will have a profession based on his age that can be: School Student, University Student, a Worker, or none of them. The percentages of people that go to University and School according to the age range are based on the CAS for Education in Lebanon [34], and the percentage of workers having different age ranges is also taken from the CAS for Labor in Lebanon [35]. These findings are summarized in *Table 3*.

Note that a worker can work in a restaurant, a nightclub, a market, a mall, a hospital, a company, a school, or a university.

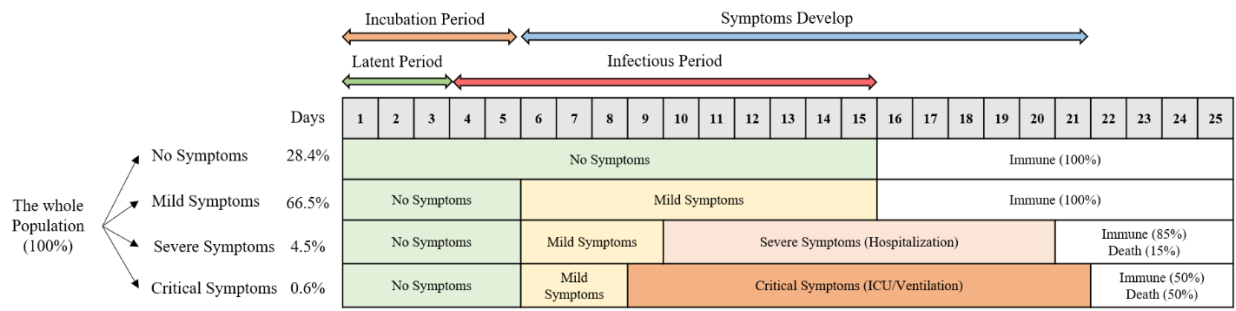

Fig. 1. Evolution of Symptoms through days, Immunity and Death percentages, and Percentage of Lebanese population that develop specific kind of Symptoms.

Table 3. Agent's Profession Distribution according to different age groups

| Age Group | School | University | Work |
|---|---|---|---|
| 0 - 4 | 26% | 0% | 0% |
| 5 - 9 | 92% | 0% | 0% |
| 10 - 19 | 71% | 9% | 7% |
| 20 - 29 | 2% | 15% | 52% |
| 30 - 39 | 0% | 0% | 62% |
| 40 - 49 | 0% | 0% | 57% |
| 50 - 59 | 0% | 0% | 47% |
| 60 - 69 | 0% | 0% | 31% |
| 70 - 79 | 0% | 0% | 16% |

*Locations to Visit:* Besides his work and study locations, an agent can go to random locations like restaurants, markets, companies, hospitals, etc. In our country, the percentages of people that go to such locations are assumed and summarized in *Table 4*.

One thing to note is that the study location and the work location for an agent are fixed, which means he has to go there every day (or at least when they are open). But in other locations, he can visit them during random times where he has nothing to do. For example, a University student does not have to go to the market every single day but he should go to the university each day during the week.

Table 4. Distribution of Locations that an agent can visit randomly, other than his work or study locations

| Age Group | Churches Mosques | Restaurants Night Clubs | Markets Malls | Hospitals | Companies |
|---|---|---|---|---|---|
| 0 - 4 | 0% | 0% | 0% | 2% | 0% |
| 5 - 9 | 0% | 4% | 16% | 2% | 0% |
| 10 - 19 | 20% | 6% | 16% | 2% | 10% |
| 20 - 29 | 20% | 8% | 6% | 2% | 7% |
| 30 - 39 | 20% | 12% | 24% | 2% | 8% |
| 40 - 49 | 20% | 10% | 6% | 2% | 6% |
| 50 - 59 | 20% | 2% | 12% | 2% | 7% |
| 60 - 69 | 20% | 0% | 6% | 4% | 6% |
| 70 - 79 | 0% | 0% | 0% | 4% | 5% |
| 80 - 89 | 0% | 0% | 0% | 4% | 0% |

*House:* Each agent has his own house where he stays if he does not have other occupations.

*Religion:* Each agent has his own religion. This religion is a characteristic of the house where an agent is born, and hence, it will be the same for all the family members.

*Houses to Visit:* For the simulation to take a realistic Lebanese aspect, we give some of the agents the ability to make visits to other houses. In our simulator, an agent can visit 2 houses: one for his friend, and one for his relatives. A visit lasts for 2 hours, not more. One thing to note is that not all agents do visits. This is mainly a characteristic that an agent can have once created.

*Travel Characteristics:* We also give the ability for some of the agents to travel in and out of the country. In order for an agent to travel, there are some conditions that we will talk about in the Simulation Process Section. During the travel days, an agent cannot interact with agents inside of the country.

## V- Building the Population

The process of building the population happens according to a specific flow. First thing, the Locations are built.

Among these locations, Houses are the origins of the population. This is where an agent is born. As discussed before, in the country structure, there is a specific number of houses that the user chooses. We use Monte Carlo Algorithm to decide the number of people inside of a house according to the House Population Distribution. After knowing how many people live in the house, we need to know their ages. Again, we use Monte Carlo to decide the age for each of them using the Age Distribution. Each house has a specific Religion, and hence, each of its members has the same religion. The religion of a house is determined via Monte Carlo also according to the Religion distribution inside of the country.

In our country, we built 15,000 houses which gave us 59,809 agents.

Now that everything is set, we have to understand the simulation process.

## VI- Simulation Process

The simulation starts with one agent being infected, and the others are susceptible (can be infected). The infected agent will go into the world, interact with other agents, infect them, etc. This is how the number of infected agents grows. The simulation process runs for several days. At the beginning of each day, an agent is assigned a list of locations that he will be at during each hour of the day (like a schedule). This list is given based on the profession of the agent (it should include his fixed locations), and on the locations where he can possibly go that are sampled using Monte Carlo Algorithm from *Table 4*. This is how agents meet at different locations.

Note that when an agent is present in a specific location, he is considered to be in a room inside of this location. This is why different locations are modeled as different rooms, each of them having specific characteristics, and based on that, each location has a defined infection rate. This rate indicates the chance of an infected agent infecting another agent within the same location (or room) during a one-hour period. In our study, we used a COVID-19 risk calculator developed by the Harvard School of Public Health [36] that is based on their peer-reviewed paper [37]. This calculator helped us approximate the risk of infection for different kinds of rooms where an agent might be. This risk percentage takes into consideration a lot of factors that include the space of the room, the activity that is done inside of it, the precautions taken by the agents inside of it, the quality of the HVAC system, the natural air condition in the room, the time spent in the room, the fact that people inside of the room are wearing masks or not, and what kind of masks they are wearing, the amount of social distancing in the room, etc. So, each of our locations was modeled as a room that has specific characteristics, and based on that we got the infection rates that are summarized in *Table 5*. Note that these rates are for one hour inside of the location.

An agent is capable of infecting others only during the infectious period that lies between the latent period, and 10 days after the incubation period [26]. Once an agent becomes infected, he will develop symptoms. If these symptoms require hospitalization (Severe, and Critical) (*Fig. 1*), we assume that the agent will be locked in a hospital and will not be able to infect others because people who visit him will be extra protective.

In addition to the locations where an agent can go, the agent can make visits (which reflects our Lebanese Cultures). For this purpose, each agent has two houses he can visit. One for his parents or relatives, and one for a friend.

An agent can also travel during the simulation. Mainly, in real cases, if the airport is open, the agent can travel only if he did not test positive in his PCR test. So, we brought this to the simulation by limiting the travel only to people who do not test positive. In other words, for an agent to travel, he has to be not infected, or infected but still in the incubation period.

An agent can stay in his travels between 3 to 5 days (this is decided according to Monte Carlo).

When an agent comes back to the country, he can either be infected or not. If he is infected, we assume that he is during the incubation period also (since he has not traveled more than this period).

At any time during the simulation, we can introduce specific measures to reduce the severity of the pandemic.

These measures can be forcing people to wear masks, or closing some locations for a specific number of days, closing the airport, etc. The effects of these actions can be monitored from real-time charts through the simulator and can give an insight into the effectiveness of a control measure at a specific time.

The agent's behavior during each day is summarized by the flowchart represented in *Fig. 2*.

## VII- Results

### A. Validating the Simulator on COVID-19 Concepts

To make sure the simulator is functioning correctly, we test two theoretical concepts that we hear a lot when it comes to COVID-19, which are: Flattening the curve, and the Second Wave Scenario.

#### 1. Flattening the Curve

Since the beginning of the pandemic, we all hear the expression "Flattening the curve" a lot [38]. This expression originates from the Centre for Disease Control and Prevention [39], and it means reducing the number of people that are simultaneously infected to avoid overwhelming the hospital care system, and thus, decrease the "peak" of currently infected people [40]. This is mainly done by taking specific measures (NPIs) in cities and countries like closing schools for example [41]. To see the effect of NPIs on the epidemic curve in our simulator, we run the following four scenarios, each of them, for 140 days. When we introduce a control measure, we start applying it on the fifth day. So, during the first five days for all of the scenarios, no control measures were taken.

Scenario 1: We do not take any control measures. All the agents can go anywhere, interact with each other, without any restrictions or precautions taken.

Table 5. Infection Rate during one hour in different locations

|  | School | University | House | Church Mosque | Restaurant | Nightclub | Market Mall | Hospital | Company |
|---|---|---|---|---|---|---|---|---|---|
| **No Mask** | 3% | 4% | 14% | 21% | 32% | 42% | 16% | 2% | 21% |
| **Mask** | 1% | 1% | 4% | 1% | 1% | 3% | 1% | 1% | 1% |

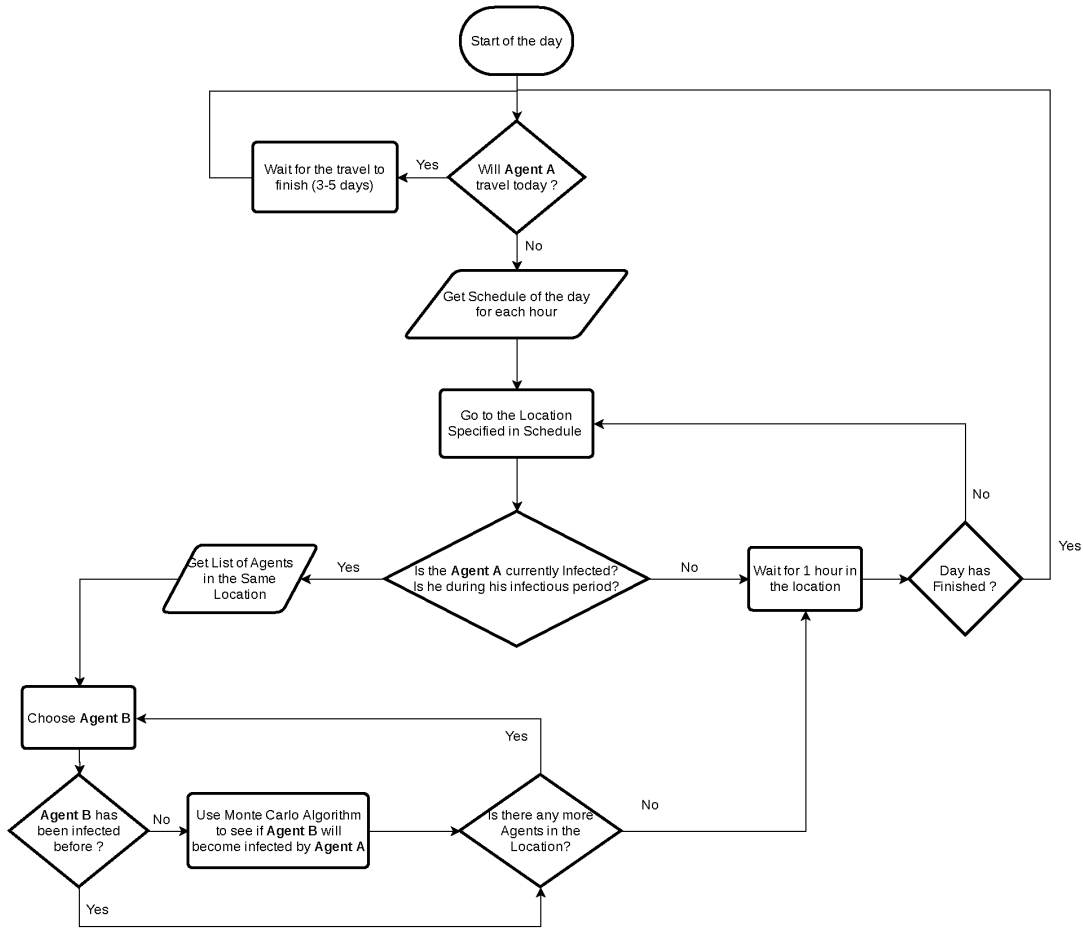

Fig. 2. Flowchart Representing a typical agent's behavior.

At the beginning of the day, we check if the Agent A will travel or not (this includes all the necessary conditions that allow an agent to travel e.g. negative PCR). If he will travel, he stays abroad for a specific duration (3-5 days). If not, he gets a schedule for the day. Each hour, the Agent A goes to a location according to this schedule. If the Agent is not infected, or is not in his infectious period, he cannot infect other agents, he stays in the location for one hour and moves to the next location or stays in the current one (according to the schedule). However, if the Agent A is currently infected and in his infectious period, he might infect others, so for each Agent B present in the same location as Agent A, if B has not been infected before, we use Monte Carlo Algorithm (based on the infection rate in this location) to determine if he will be infected by Agent A or not.

Scenario 2: We do not close any location in the country, but we force the agents to wear masks and this happens by reducing the infection rate at different locations as cited in *Table 5*.

Scenario 3: We close the airport, and force agents to wear masks.

Scenario 4: We close the airport, schools, Universities, and force agents to wear masks.

For the four scenarios, we plot the curve of the currently infected agents through time *Fig. 3*. We can see in the first scenario, the peak is the highest, and the pandemic lasts the shortest time. We can see that while introducing NPIs, the curves become flattened which means their peak decreases well, and the pandemic takes more time to end. These results validate the theoretical concept of flattening the curve in our country.

One thing to note is that through flattening the curve, we are not only limiting the number of active cases inside of a country but also, we're limiting the number of deaths because, like this, a lot of people would not get infected. This can be understood from *Fig. 4*. where we see the evolution of deaths through time for each of the four presented scenarios.

### 2. Second Wave Scenario

Also, since the beginning of the pandemic, we hear that if we open the countries too soon, we will risk a second wave of COVID-19 that may be harsher than the first one [42, 43]. So, to validate this concept, we took the following scenario:

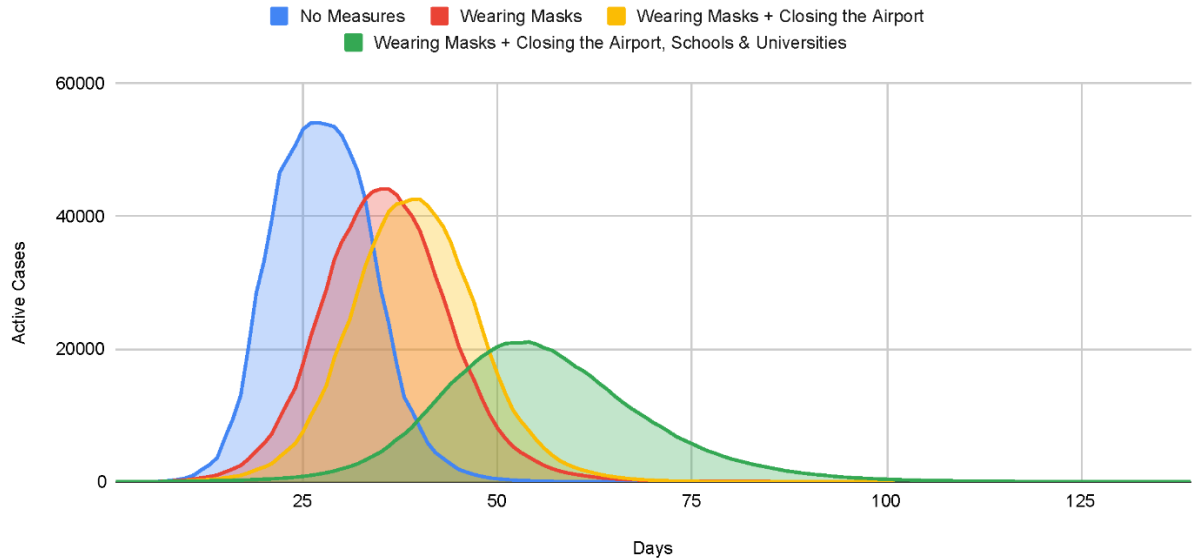

Fig. 3. Flattening the Curve through different NPIs.

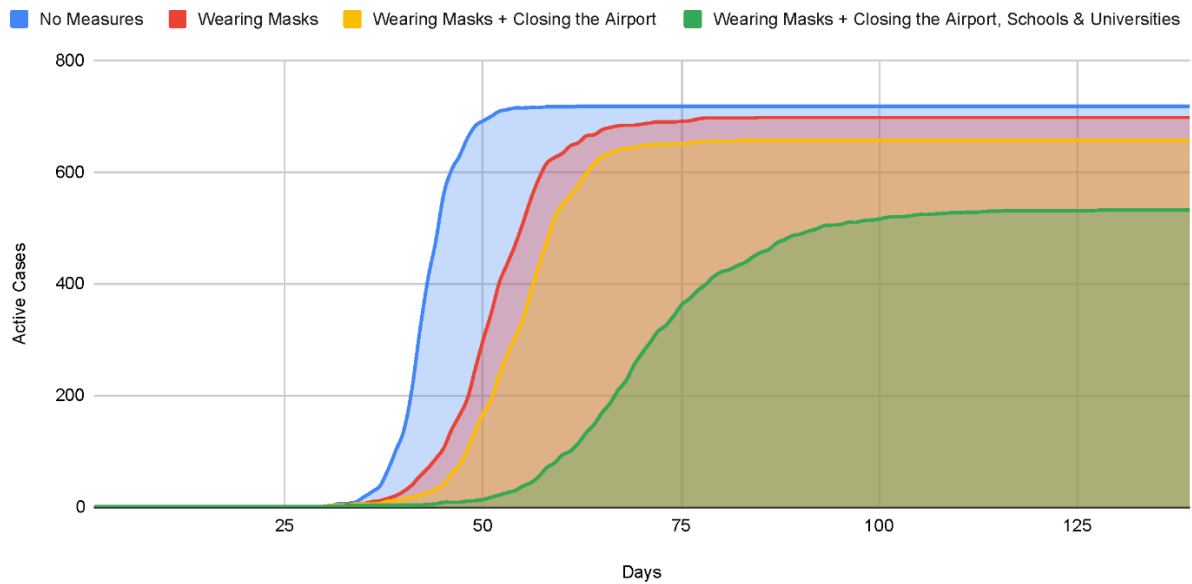

Fig. 4. Evolution of Deaths for different applied NPIs.

We introduce a Full Lockdown on day 10 (Everything is closed, we only keep hospitals open). Then on day 100 (after three months), we see that the number of currently infected people has reduced, so we lift this lockdown by allowing restaurants and nightclubs to open up again while maintaining the mask-wearing and social distancing strategy.
We plot the curve of the currently infected people through time. The results appear in *Fig. 5*. We can easily see that if we start opening closed locations and lifting the lockdown while the number of infected people is still not close to zero,
we will have a second wave, where a lot more people can get infected.

### B. Validating the Simulator on Lebanon Case

To be able to use our simulator to predict future trends in the Lebanese pandemic situation, we have to make sure that it is working well on the previous scenarios that were applied in Lebanon.
For this purpose, we simulated all the actions that have been taken in Lebanon since the start of the pandemic until 21 April 2021, to see their effect on our virtual country, and then

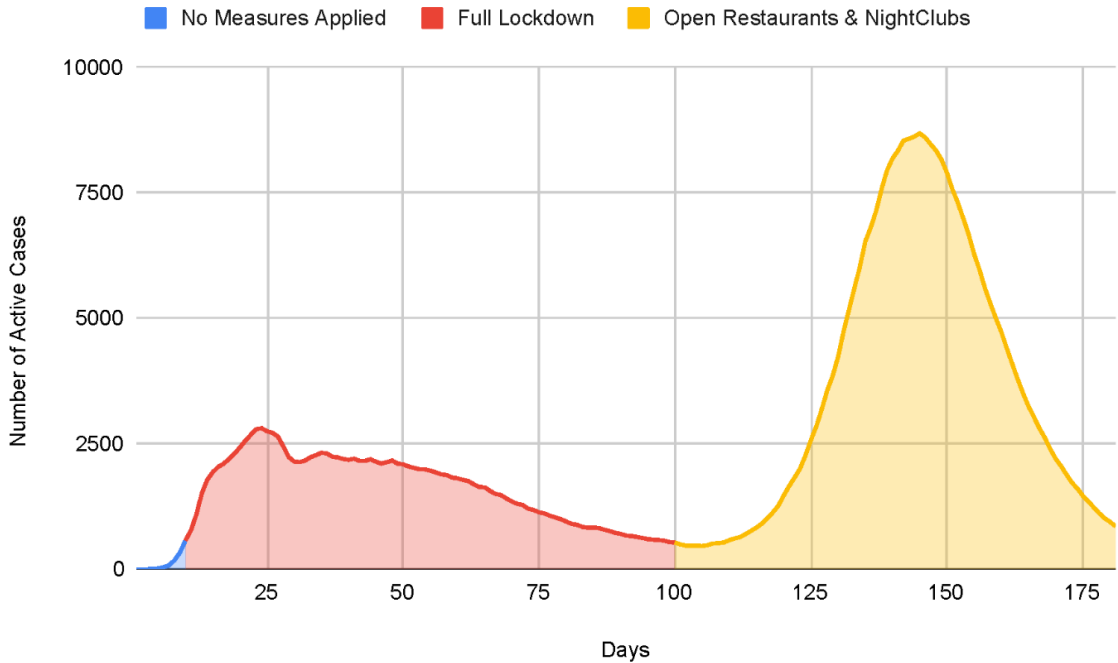

Fig. 5. Second Wave Scenario.

compare our "Active cases" pandemic curve with the real one of Lebanon.

The difference between our country and Lebanon is simply the number of agents. Our country has around 60,000 agents, but Lebanon has around 6,801,083 people living inside of it [44]. So, the duration of the pandemic and the dynamicity of the virus spread will differ of course. For this reason, we apply the control measures, not at their exact times as Lebanon, but we apply all of them in chronological order to obtain a curve similar to the real active cases curve available on the Worldometer's page [45].

The information about measures that were applied in Lebanon is taken from the Disaster Risk Management Unit [46]. In Lebanon, the first case was registered on February 21, 2020 [47]. After that by six days, daycare centers, schools, and universities were shut [41]. On March 6, 2020 Gyms, pubs, theaters, and nightclubs were closed [47]. After that (March 11, 2020), all malls, and restaurants were closed. On March 15, 2020, the Government initiated a full lockdown and even closed the airport. On 27 April 2020, a phased reopening started and ended on 1 July 2020 with the airport reopening. After that, on 27 July 2020, the government re-initiated Lockdown. This lockdown lasted till August 10, 2020. After that, by the end of the summer, schools started reopening at half capacity. The number of infections started rising again, and this is why on 14 November 2020, the lockdown was initiated again. It was lifted on 30 November 2020. And then finally, on 7 January 2021, a new lockdown was initiated and lasted till 8 February 2021, when the phased reopening started.

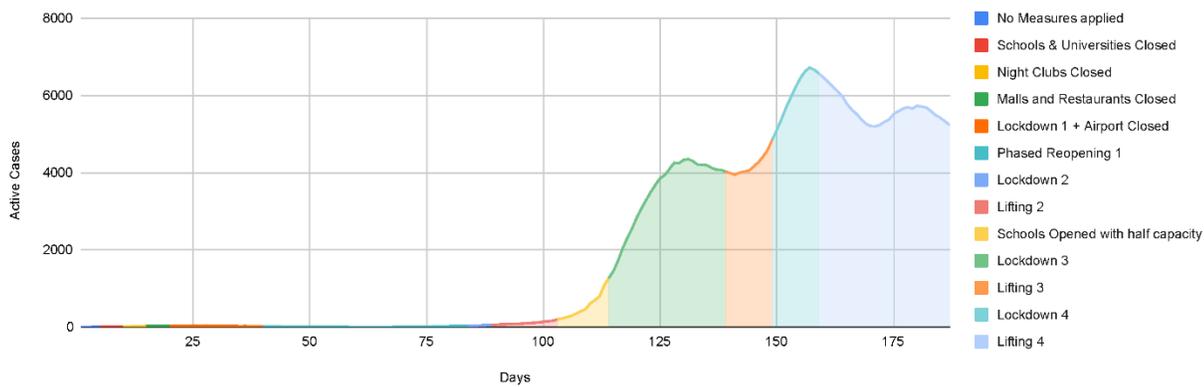

Fig. 6. Simulated Active Cases in Lebanon.

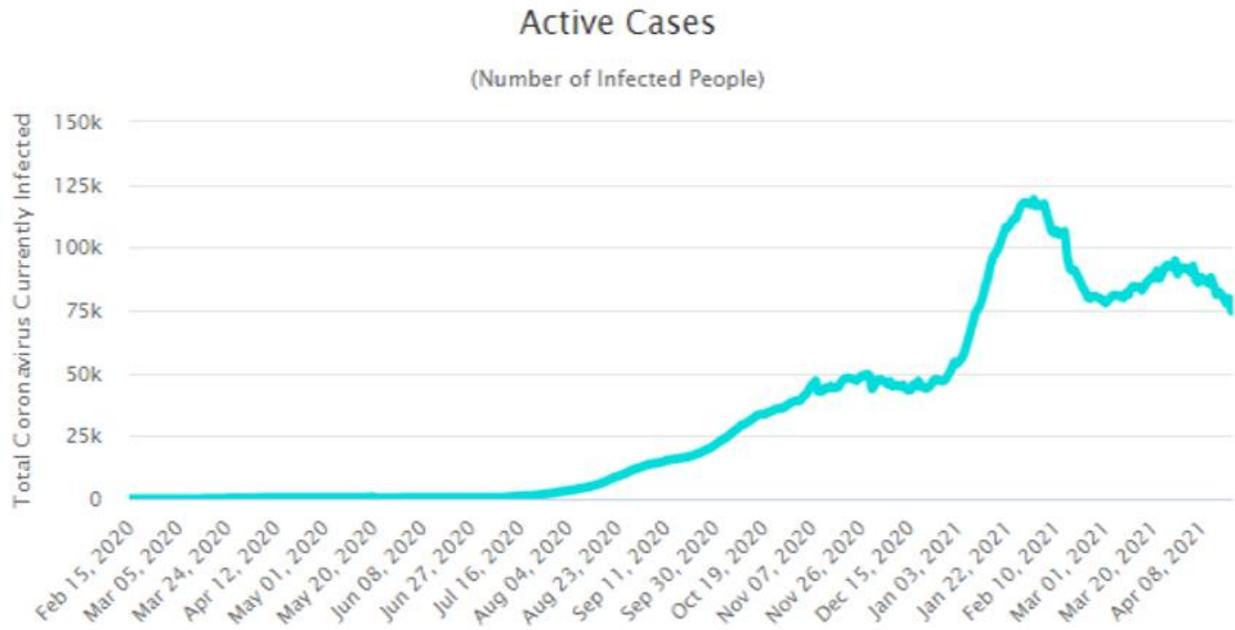

Fig. 7. Real Active Cases Curve in Lebanon – Source: Worldometer Lebanon's Page [45].

We applied the same control measures in our simulation but at different times as clarified in *Fig. 6*, and we end up with a very similar curve to the Lebanon active cases available on the Worldometer's page shown in *Fig. 7* [45].

## VIII-Forecasting based on Future Scenarios

Now that we have a curve that represents the Lebanese pandemic situation, we can predict what will happen in the future based on our actions. In fact, the Lebanese Ministry of Education expressed their willingness to open schools again progressively. This is why we wanted to see the effect of such action via our simulator, especially that the vaccination process is still slow in the country.

For this purpose, we ran the following 4 scenarios:

Scenario 1: We do not open schools and universities.
Scenario 2: We open schools alone
Scenario 3: We open Universities alone
Scenario 4: We open Schools and Universities at the same time

We can see the result of each of those actions in *Fig. 8*. It is clear that opening now especially with the delays in the vaccination campaign would be harmful for the pandemic situation inside the country. In addition to that, the number of deaths will rise more sharply after such actions are applied according to our simulations in *Fig. 9*. This is why such an action should be postponed until the vaccination campaign progresses more effectively, especially that until 26 April 2021, we only had 4% of the whole population who have taken the first dose of the vaccine, and only 2.2% who were fully vaccinated [16].

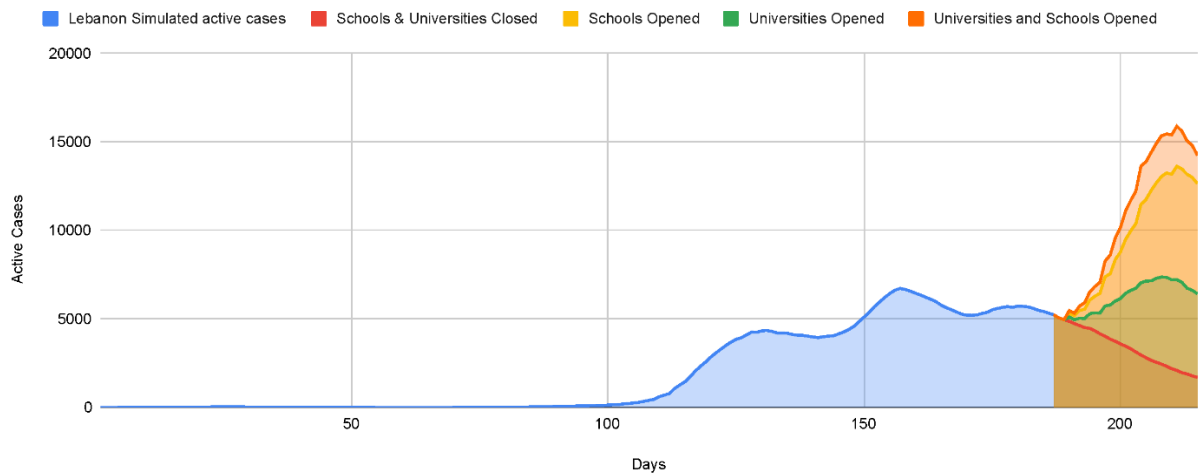

Fig. 8. Predicted Evolution of Active Cases based on different taken actions.

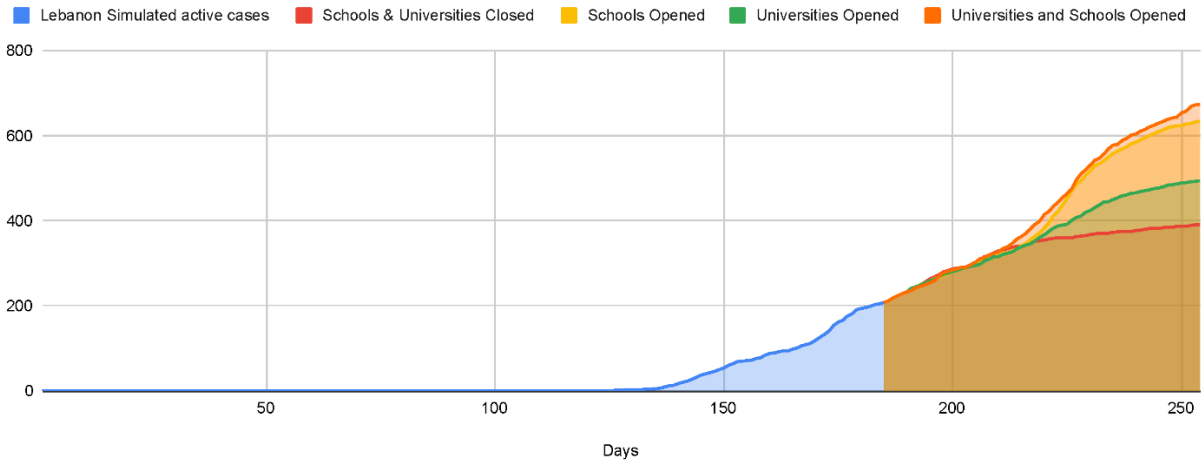

Fig. 9. Predicted Evolution of Deaths based on different taken actions.

## IX- Conclusion

It is clear that opening schools or universities or both will have a negative effect on the pandemic situation in the country, as we will experience an increasing number of infections that can overshoot the maximum that we have reached so far. This is why such a decision should be procrastinated a bit till we have a good vaccination process running inside of the country.

## X- Improvements and Future Works

A lot of things can still be added to our simulator. 1) The locations where an agent might go can be taken from real people instead of them being based on Monte Carlo Algorithm applied to statistical distributions. 2) Finding the optimal strategy to follow in a country to reduce the impacts of the virus. 3) The vaccination process and the vaccination strategy that should be followed to get the best outcome and limit the spread of the virus at the earliest time possible.

## XI- Acknowledgements

This work has been financially supported by the Lebanese University. We would like to acknowledge their efforts.

## XII- References